\documentclass[journal]{IEEEtran}

\usepackage{xcolor,soul,framed} 

\colorlet{shadecolor}{yellow}
\usepackage[pdftex]{graphicx}
\graphicspath{{../pdf/}{../jpeg/}}
\DeclareGraphicsExtensions{.pdf,.jpeg,.png}

\usepackage[cmex10]{amsmath}
\usepackage{array}
\usepackage{mdwmath}
\usepackage{mdwtab}
\usepackage{eqparbox}
\usepackage{url}

\usepackage{epsfig}
\usepackage{graphicx}
\usepackage{amsmath}
\usepackage{amssymb}
\usepackage{multirow}
\usepackage{booktabs}
\def\blu#1{\textbf{\color{blue} #1}} 
\def\red#1{\textbf{\color{red}  #1}} 
\usepackage{amsmath}
\usepackage{tabularx}
\usepackage[pagebackref=False,breaklinks=true,letterpaper=True,colorlinks,linkcolor=black,citecolor=black,bookmarks=false]{hyperref}
\usepackage[misc]{ifsym}

\begin{document}
\bstctlcite{IEEEexample:BSTcontrol}
    \title{Progressively Dual Prior Guided Few-shot Semantic Segmentation}
  \author{Qinglong Cao,
      Yuntian Chen,~\IEEEmembership{Member,~IEEE,}\\
      Xiwen Yao,~\IEEEmembership{ Member,~IEEE,}
      and~Junwei~Han,~\IEEEmembership{Fellow,~IEEE}

  \thanks{This work was supported in part by the National Science Foundation of China under Grants 62071388, 62136007, the Key R\&D Program of Shaanxi Province under Grant 2021ZDLGY01-08}
  \thanks{Q. Cao is with MoE Key Lab of Artificial Intelligence, AI Institute, Shanghai Jiao Tong University, Shanghai, 200240, China and Eastern Institute for Advanced Study, Zhejiang, China (e-mail: caoql2022@sjtu.edu.cn).}
  \thanks{Y. Chen is with Eastern Institute for Advanced Study, Zhejiang, China (e-mail: ychen@eias.ac.cn).}%
  \thanks{X. Yao and J. Han are with the School of Automation, Northwestern Polytechnical University, Xi’an 710072, China (e-mail: yaoxiwen517@gmail.com,junweihan2010@gmail.com ).}
  \thanks{\textit{Corresponding author: Xiwen Yao}}
}

\markboth{Journal of \LaTeX\ Class Files,~Vol.~14, No.~8, August~2015}
{Shell \MakeLowercase{\textit{et al.}}: Bare Demo of IEEEtran.cls for IEEE Transactions on Magnetics Journals}

\maketitle

\begin{abstract}
Few-shot semantic segmentation task aims at performing segmentation in query images with a few annotated support samples. Currently, few-shot segmentation methods mainly focus on leveraging foreground information without fully utilizing the rich background information, which could result in wrong activation of foreground-like background regions with the inadaptability to dramatic scene changes of support-query image pairs. Meanwhile, the lack of detail mining mechanism could cause coarse parsing results without some semantic components or edge areas since prototypes have limited ability to cope with large object appearance variance. To tackle these problems, we propose a progressively dual prior guided few-shot semantic segmentation network. Specifically, a dual prior mask generation (DPMG) module is firstly designed to suppress the wrong activation in foreground-background comparison manner by regarding background as assisted refinement information. With dual prior masks refining the location of foreground area, we further propose a progressive semantic detail enrichment (PSDE) module which forces the parsing model to capture the hidden semantic details by iteratively erasing the high-confidence foreground region and activating details in the rest region with a hierarchical structure. The collaboration of DPMG and PSDE formulates a novel few-shot segmentation network that can be learned in an end-to-end manner. Comprehensive experiments on PASCAL-$\textit{5}^\textit{i}$ and MS COCO powerfully demonstrate that our proposed algorithm achieves the great performance.
\end{abstract}


%
\IEEEpeerreviewmaketitle


\section{Introduction}
\IEEEPARstart{D}{eep} learning technology has achieved breakthrough improvements in many visual tasks: image classification~\cite{A1,A2,A3,A59}, object detection ~\cite{A4,A5,A6,A58}, and semantic segmentation~\cite{A7,A8,A9,A57}. Most state-of-the-art algorithms are based on large scale labeled datasets, such as Imagenet ~\cite{A10} and PASCAL VOC dataset\cite{A35}. However, to acquire enough annotated datasets is laborious and expensive, especially in semantic segmentation task. Meanwhile, it is impossible to get abundant annotated datasets in diverse application areas in our real world. Subsequently, few-shot semantic segmentation task is proposed to tackle this challenge, where datasets are divided into base class set and novel class set.

\begin{figure}[t]
	\begin{center}
		\includegraphics[width=1.0\linewidth]{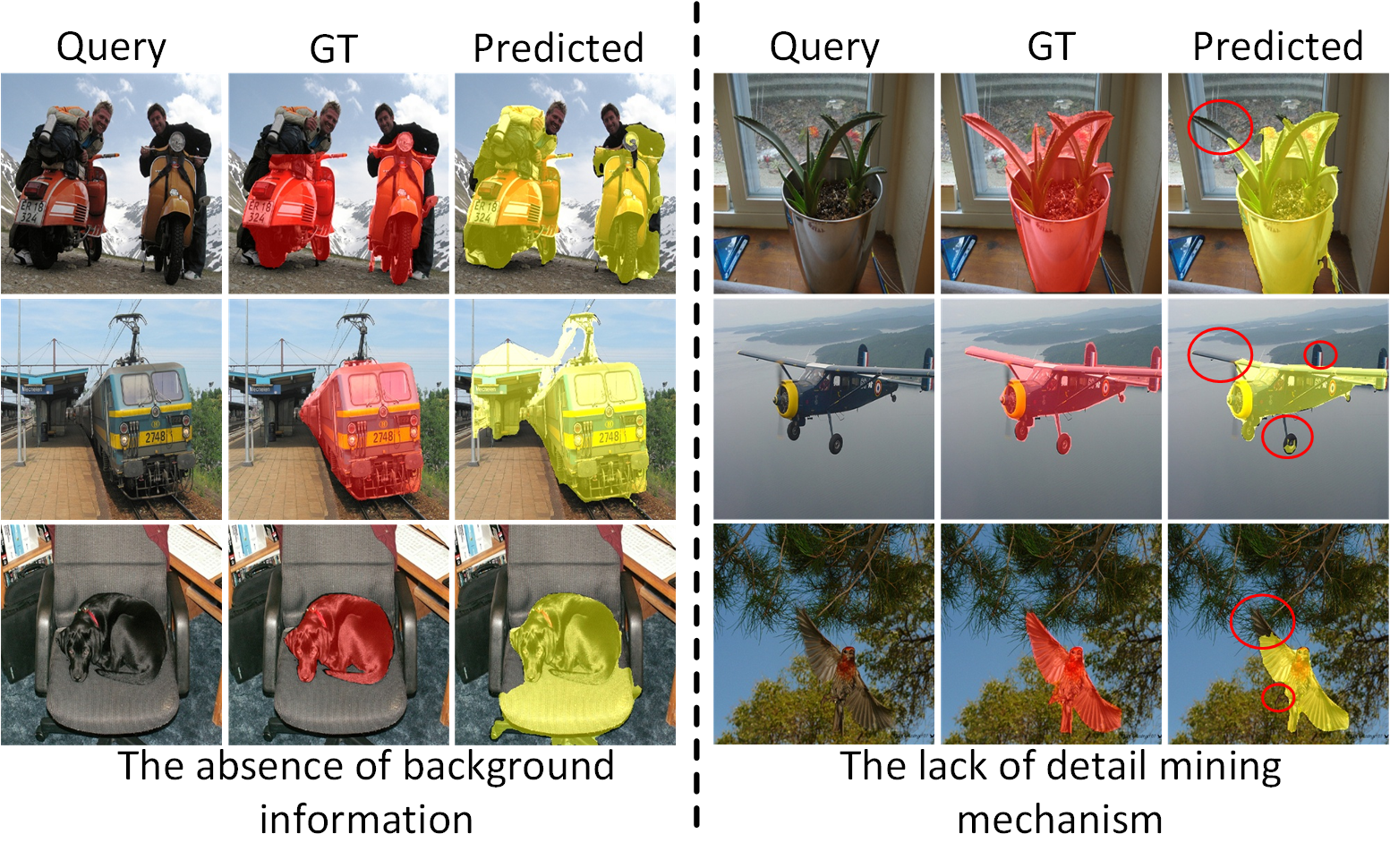}
	\end{center}
	\caption{Overview of existing drawbacks for current few-shot segmentation algorithms. Background is neglected in most few-shot segmentation algorithms, which results in the activation of wrong regions. Meanwhile, the lack of detail mining mechanism leads to coarse results with some semantic object components or edges missing.}
	\label{fig:first}
\end{figure}

Most previous few-shot segmentation algorithms generally feed the support images and query images into a weight-shared module, and learn support prototypes from support samples. Then the query images would be accurately parsed by elaborately processing query images features that combined with support prototypes. Most existing methods generally tackle the task in this pattern by diverse angles like more representational prototypes~\cite{A11,A12,A13,A63} or better feature extracting~\cite{A14,A15,A62}.

Although remarkable progresses have been achieved in the aforementioned algorithms, there are still some challenges. Firstly, previous few-shot segmentation models mainly focus on utilizing the support foreground information to straightly discriminate the semantic objects, and the rich background regions providing functional scene information are irrationally neglected. However, suffered from the dramatic scene changes of support-query pairs, some background regions of query images could be similar in color or shape to the semantic objects in support samples. This phenomenon would naturally lead to the wrong activation of some background regions in predicted results. The second problem is the lack of detail mining mechanism in parsing process. Current few-shot segmentation methods tend to extract global support information as guiding prototypes without considering object spatial information like object postures and complicate structures. Thus the network only has ability to recognize coarse structure of foreground, which lacks some essential object components and detailed foreground edge areas.

To tackle the first challenge, we construct a dual prior mask generating (DPMG) module to produce the dual prior masks where two masks respectively expose the background and foreground possibility of query images in pixel-level, and the comparison between foreground and background probability masks helps the parsing model to exclude the foreground-like background regions. More specifically, we firstly extract the class-level informative deep feature as the input feature and leverage the labeled support mask to filter the support features as separated foreground and background features. Moreover, we would mine the relation between query and support by computing the pixel-level similarity between query features and support features, and selecting the highest activation points as corresponding values. In this way, we would generate the foreground prior mask and the opposite background prior mask, which are further concatenated to produce the dual prior masks. Each pixel of dual prior masks has two comparable probability values of foreground and background to refine the location of foreground.

With the informative dual prior masks constraining the activation of wrong background area, the progressive semantic detail enrichment (PSDE) module is further proposed to acquire more accurate results in a hierarchical manner aiming at fully utilizing various information from different levels. The query features are firstly fused with support prototypes and further concatenated with dual prior masks at each level to generate the class-aware features. The high-level class-aware features are directly fed into a series of convolutional blocks to compute the initial coarse parsing results. The initial coarse parsing results are used as erasing masks to erase the high-confidence foreground information from the lower-layer class-aware features. Subsequently, we leverage a series of convolutional blocks to activate the hidden semantic details of the lower layer class-aware features. Then the activated semantic details are corresponding added with the initial coarse result to acquire the more complete predicted results. Moreover, the more complete results are regarded as the initial coarse results for the following lower layer. By iteratively repeating the detail enrichment process between the neighbor layers, we finally get the most precise result in high resolution in lowest layer.

The cooperation of DPMG and PSDE formulates a novel few-shot segmentation network where the dual prior masks are elaborately introduced to get better foreground location and the continuous foreground erasing and detail enrichment endeavor to predict more complete semantic objects. Extensive experiments on PASCAL-$5^i$ and MS COCO datasets clearly demonstrate the superiority of our network. The main contributions are summarized as follows:
\begin{itemize}
	\item The DPMG module is designed to refine the location of foreground by computing foreground-background similarities. Benefiting from the DPMG module, our network suppresses the activation of wrong background area with a better adaptation to dramatic scene changes among query-support images.
	
	\item The PSDE module is proposed to progressively optimize the parsing result by iteratively erasing the high-confidence foreground regions and activating the hidden semantic details with further combination of dual prior masks in a hierarchical manner. 
	
	\item Experimental results demonstrate that the proposed method achieves superior performance with many classical few-shot semantic segmentation methods.

\end{itemize}
\begin{figure*}
	\begin{center}
		\includegraphics[width=0.80\linewidth]{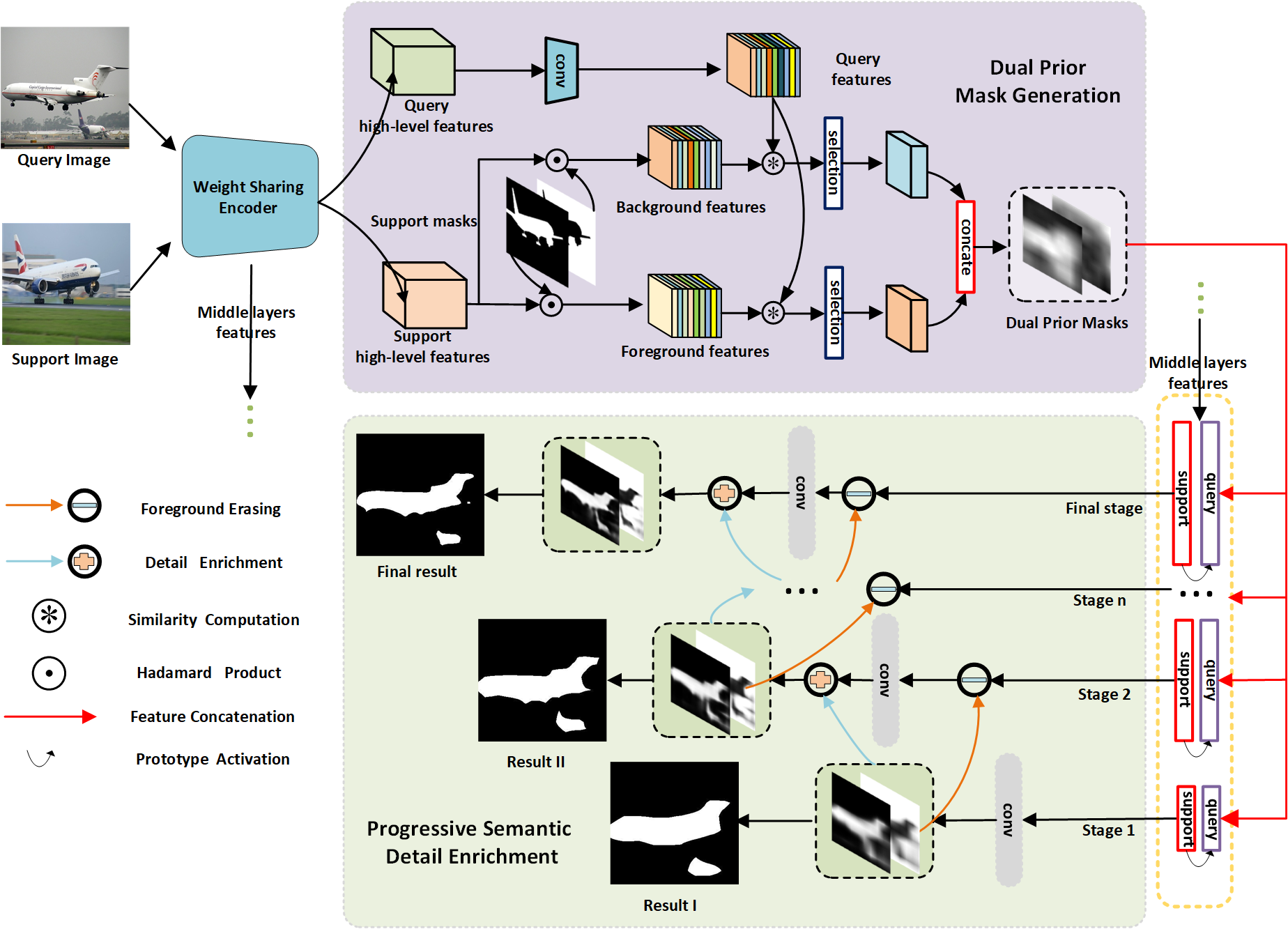}
	\end{center}
	\caption{The pipeline of our network architecture. Our network would firstly adopt high-level features to implement the dual prior mask generation module, which could refine the location of foreground. With the features from diverse levels fused with prototypes, we could utilize the class-aware features to achieve the progressive semantic detail enrichment in hierarchical manner and the prior masks could guide the process to acquire accurate predicted results.}
	\label{fig:main}
\end{figure*}

\section{Related Work}

\subsection{Semantic segmentation.} 
semantic segmentation is a fundamental topic aiming at predicting semantic class of each pixel. Most current methods are based on deep convolutional neural network. Long \textit{et al.} ~\cite{A8} firstly proposes Fully Convolutional Network (FCN) which achieves significant segmentation performance improvement. Receptive field is crucial in segmentation task, thus Deeplab ~\cite{A9} and Yu \textit{et al.}~\cite{A16} utilize the dilated convolution to reinforce result. Furthermore, Zhao \textit{et al.}~\cite{A17} designs the PSPNet, which captures the global information by context aggregation. Popular Ecoder and Decoder architecture~\cite{A18} is also introduced to boost the segmentation performance. Similarly, this work~\cite{A8} designs a encoder-decoder like U-Net to perform the accurate semantic segmenation with features from different layers in a dense skip connected manner. Furthermore, based on the DeepLabV3\cite{A46}, Chen \textit{et al.}~\cite{A47} construct the DeepLabV3-Plus by connecting the encoder with the dense decoder components.

Some works \cite{A48,A49,A50,A51} tend to utilize the deconvolution in the parsing process to ehance the semantic segmentation performance.  Noh \textit{et al.}  ~\cite{A48} firstly proposes the deconvolution-based DeconvNet to perform the semantic segmentation task. Based on the DeconvNet~\cite{A48}, Fourure \textit{et al.}~\cite{A49} further proposed the GridNet, of which the deconvolution is utilized in the decoder component to recover spatial resolution. Meanwhile, this work ~\cite{A50}incorporate the bayesian knowledge in the semantic segmentation task to consturct the accurate Bayesian-SegNet.

To further reinforce the semantic segmentation performance, many works attempt~\cite{A39} to mine the hidden sequential semantic information with the recurrent neural network (RNN). For instance, RCNN~\cite{A39} would predict several coarse predictions with different plain convolutional blocks and further obtian the final parsing results in complex refining process. Aiming at lower the huge computational cost in the RCNN~\cite{A39}, this work~\cite{A40} separates original images into diverse non-ovelapping patches as the inputs of parsing model. Meanwhile, Visin \textit{et al.}~\cite{A43} proposes the ReSeg method to capture the long-range dependencies of semantic objects based on the local feature. To address the dependencies fading challenge during the long-term processing, Fan \textit{et al.}~\cite{A42} proposes the Dense Recurrent Neural Network to build the dense connections between each patch of the images.

Focusing on obtaining high-level semantic features while maintaining spatial information, this work ~\cite{A44} proposes a double-substream based network, of which the pooling stream aims at extract the high-level features and the residual stream attempts to capture the spatial details. Recently, DANet~\cite{A45} adopts the dual position and channel attention to help the network further mine the spatial and channel interdependencies. Based on the dual attention mechanism, Liu  \textit{et al.}~\cite{A43} further proposes the polarized self-attetion to improve the segmentation performance with better pixel-wise regression. All previous algorithms  decrease the initial image resolution during the parsing proces, which would obviously hamper the segmentation performance. To conquer it, the RefineNet~\cite{A7} proposes a generic multi-path refinement network to extract the complete information in segmentation. Inspired by the hierarchical architecture, we design a likewise multi-path architecture to mine the semantic details with neglected background information introducing. 

\begin{figure*}
	\begin{center}
		\includegraphics[width=0.8\linewidth]{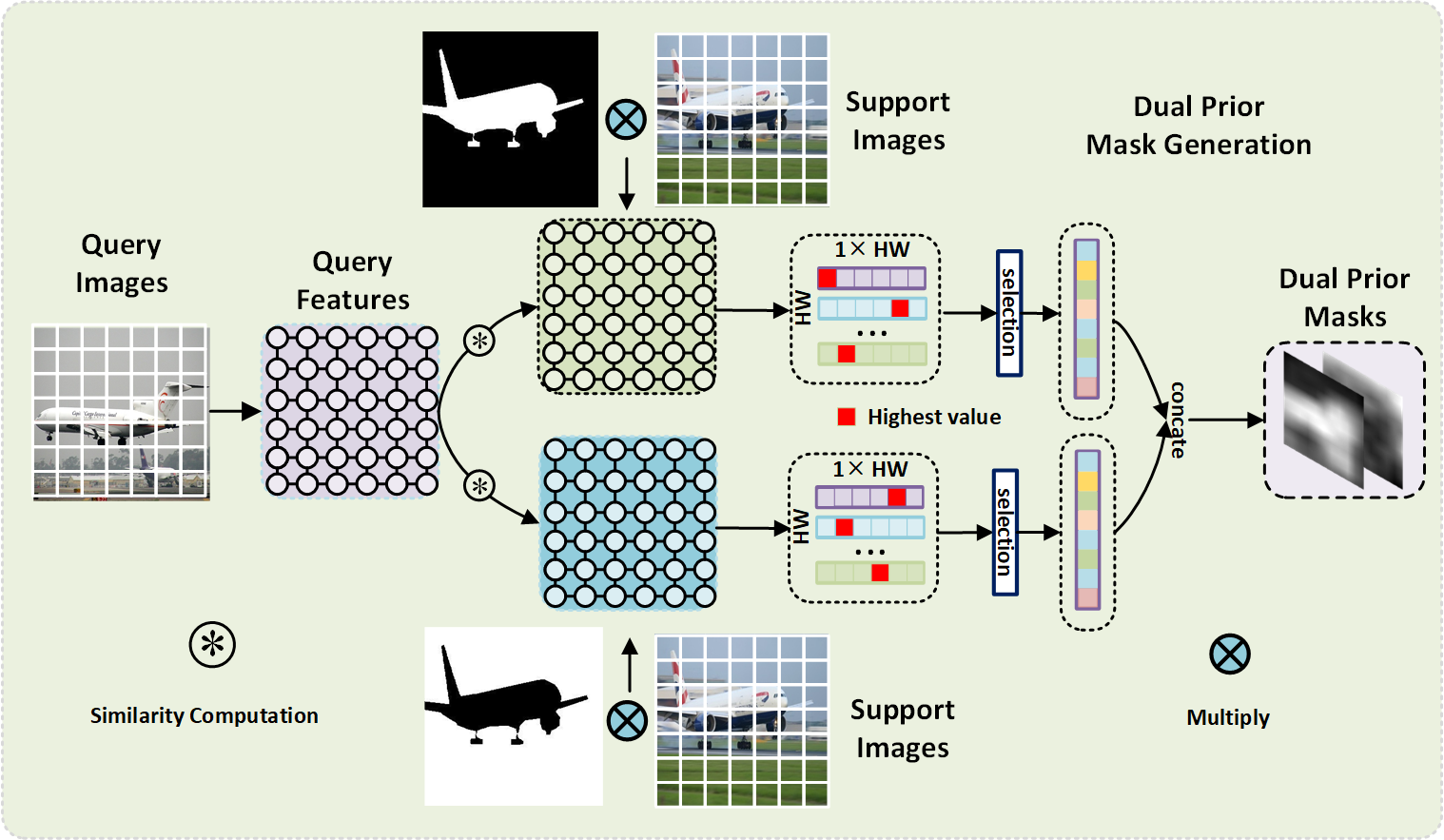}
	\end{center}
	\caption{The illustration of dual prior mask generation module. By computing the similarities between query features and support features of foreground and background, and selecting the highest activation as the corresponding value, the foreground and background prior masks would be automatically generated. Subsequently, the comparison between the two prior masks would help the network constraining the activation of wrong background areas.  }
	\label{fig:graph}
\end{figure*}

\subsection{Few-shot semantic segmentation.} 
The few-shot semantic segmentation task aims at how to accurately segment large quantities of query images based on a few annotated support images. By introducing the prototype concept into few-shot segmentation domain, Dong \textit{et al.}~\cite{A53} proposes the first generalized few-shot semantic segmentation framework in an alternative training scheme. Based on this work~\cite{A14}, the SG-one network~\cite{A11} extracts a guiding feature from support images and utilize the similarity between the prototypes and query images to achieve the segmentation task. However, the network neglects the hierarchical information, based on it, CAnet~\cite{A14} proposes a multi-level feature comparison model to optimize the segmentation result iteratively. Moreover, PANet~\cite{A12} introduces a prototype alignment regularization to get a more generalized prototype to improve the segmentation performance. Meanwhile, the PMMs~\cite{A29} designs the multiple prototypes to enhance the segmentation performance by better semantic representations. Similarly, by decomposing the holistic class representation into a set of part-aware prototypes, PPNet~\cite{A55} precisely capture diverse and fine-grained semantic object features to boost the parsing performance.  By introducing the attention mechanism into the few-shot semantic segmentation task, A-MCG net~\cite{A60} effectively perform the multi-context guiding to generate more precise parsing. Aiming at providing more accurate guidance for the segmentation process, ASGnet~\cite{A56} consturct the guided prototype allocation to assign each pixel of query images with the most similar prorotype. The knowledge transfer challenge from base class to new class is not addressed in prior mentioned algorithms. Inspired by it, The AMP network~\cite{A13} fuses the new and old knowledge to get the adaptive masked proxies feature to achieve effective segmentation. With the rising of graph knowledge in deep learning network, PGnet~\cite{A15} attempts to introduce  graph knowledge into few-shot segmentation domain by building a pyramid graph network. These methods mainly focus on the foreground similarity comparison between query and support but ignore essentiality of background. Meanwhile, the guided prototypes have limited ability to represent the complicate object structures. The two dilemmas seriously hamper the performance of segmentation models.

\section{Method}

\subsection{Problem Setting}
A few-shot semantic segmentation system generally has two subsets, the pixel-level annotated support set and the unlabeled query set, where two datasets has no overlapped categories. The goal of the system is to discover the foreground pixels in query images with a few annotated samples from support set.

Suppose we have abundant data from two non-overlapping sets of classes $C_{base}$ and $C_{novel}$. The training dataset $D_{train}$ is constructed on $C_{base}$, and the test dataset $D_{test}$ is constructed from $C_{novel}$. Given $k$-shot setting scenario, we sample $k+1$ labeled images $\left\{ (I_s^1,M_s^1),(I_s^2,M_s^2), \cdot  \cdot  \cdot (I_s^k,M_s^k),({I_q},{M_q})\right\}$ in targeted class from the  $D_{train}$ episodically, where $(I_s^i,M_s^i)$  means the support pair and $({I_q},{M_q})$  means the query pair. During training phase, the $k$-shot support pairs are inputted into the model to help the model make the prediction ${\hat M_q}$ on ${I_q}$ and the model is iteratively optimized by the cross-entropy loss $\ell ({\hat M_q},{M_q})$. During the test process, we implement the same operation to get the test images in novel classes.

\subsection{Method Overview}

As shown in Figure \ref{fig:main}, we propose a novel and effective few-shot semantic segmentation model by constraining the wrong activation of background areas and iteratively mining the hidden semantic details with hierarchical information. The overall framework mainly consists of two modules: dual prior mask generation module and progressive semantic detail enrichment module. The dual prior mask generation module is firstly constructed to produce the dual prior masks with high-level class-aware features, where two masks respectively denote the foreground and background possibility of query images. By directly comparing the values of two prior masks, we acquire a better location of foreground in query images. Based on the dual prior masks, the progressive semantic detail enrichment module is further proposed to capture the semantic details from diverse levels with a hierarchical architecture. Specifically, the foreground erasing process helps the network focus on mining details by iteratively erasing the high-confidence foreground area, and detail enrichment works on the fusion of parsing results from diverse layers.
\subsection{Dual Prior Mask Generation}
The whole architecture of dual prior masks generation module is shown in Figure \ref{fig:graph}. In order to precisely measure the relation between the query and support, we implement the pixel-level similarity computation between query and support features to produce dual prior masks. Assume we have query features ${F_q}$ and support features ${F_s}$, where the size of  ${F_q}$ and  ${F_s}$ equals $C \times H \times W$. Subsequently features are respectively flattened in spatial dimension as query features sets $\left\{ { l_1^q,l_2^q, \cdot  \cdot  \cdot , l_N^q} \right\}$ and support features sets $\left\{ {l_1^s, l_2^s, \cdot  \cdot  \cdot ,l_N^s} \right\}$, where $ l \in {C}$ and $N = H \times W$.  The support sets are filtered as two subsets by support masks: support foreground features $ l_i^f$, $i \in foreground$, and support background features $ l_j^b$, $j \in background$, where $f$ means the foreground pixels and $b$ denotes the background pixels.
\begin{equation}
\left\{ { l_1^f,l_2^f, \cdot  \cdot  \cdot , l_i^f} \right\} = {F_s} \odot {M_s}
\end{equation}

\begin{equation}
\left\{ { l_1^b,l_2^b, \cdot  \cdot  \cdot , l_j^b} \right\} = {F_s} \odot (1 - {M_s})
\end{equation}

We choose the cosine similarity as the correlation function ${f_c}\left(  \cdot  \right)$ to compute the pixel-level relation scores between query features and support features:

\begin{equation}
\cos (l_k^q,l_i^f) = \frac{{l_k^ql_i^f}}{{\left\| {l_k^q} \right\|\left\| {l_i^f} \right\|}},k \in \{ 1,2,....,N\} ,i \in foreground
\end{equation}
\begin{equation}
\cos (l_k^q,l_j^b) = \frac{{l_k^ql_j^b}}{{\left\| {l_k^q} \right\|\left\| {l_j^b} \right\|}},k \in \{ 1,2,....,N\} ,j \in background
\end{equation}
the highest activation scores in support dimension are selected as the corresponding values for each loaction of query features:
\begin{equation}
e_n^f = \mathop {\max }\limits_{i \in foreground} {f_c}(l_n^q,l_i^f)
\end{equation}
\begin{equation}
e_n^b = \mathop {\max }\limits_{j \in background} {f_c}(l_n^q,l_j^b)
\end{equation}
where $e_n^f$  is the positive similarity score of $n-th$ query pixel and $e_n^b$ is the negative similarity score of $n-th$ query pixel. The high value of $e_n^f$/$e_n^b$  in one pixel of query features would imply that at least one pixel in support foreground/background has high correlation with this query pixel. Therefore, this query pixel has fairly high possibility of belonging to foregound/background. Based on this fact, we gather the generated similarity scores to produce foreground probability map ${m_f} = \left\{ { e_1^f, e_2^f, \cdot  \cdot  \cdot , e_N^f} \right\}$ and background probability map ${m_b} = \left\{ { e_1^b, e_2^b, \cdot  \cdot  \cdot , e_N^b} \right\}$. These maps would  imply the foreground/background probability of each pixel in query images. Finally, we concatenate the two probability maps to generate the dual prior masks.
\begin{equation}
{m_s} = concate({m_f},{m_b})
\end{equation}
The dual prior masks could provide each pixel of query images with two probability scores, and the pixel-level comparison could precisely refine the location of foreground area.
In $k$-shot setting, we would generate k-shot dual prior masks and regard the mean of the k-shot masks as the final dual prior masks.

\subsection{Progressive Semantic Detail Enrichment}
\begin{figure}[t]
	\begin{center}
		\includegraphics[width=1.0\linewidth]{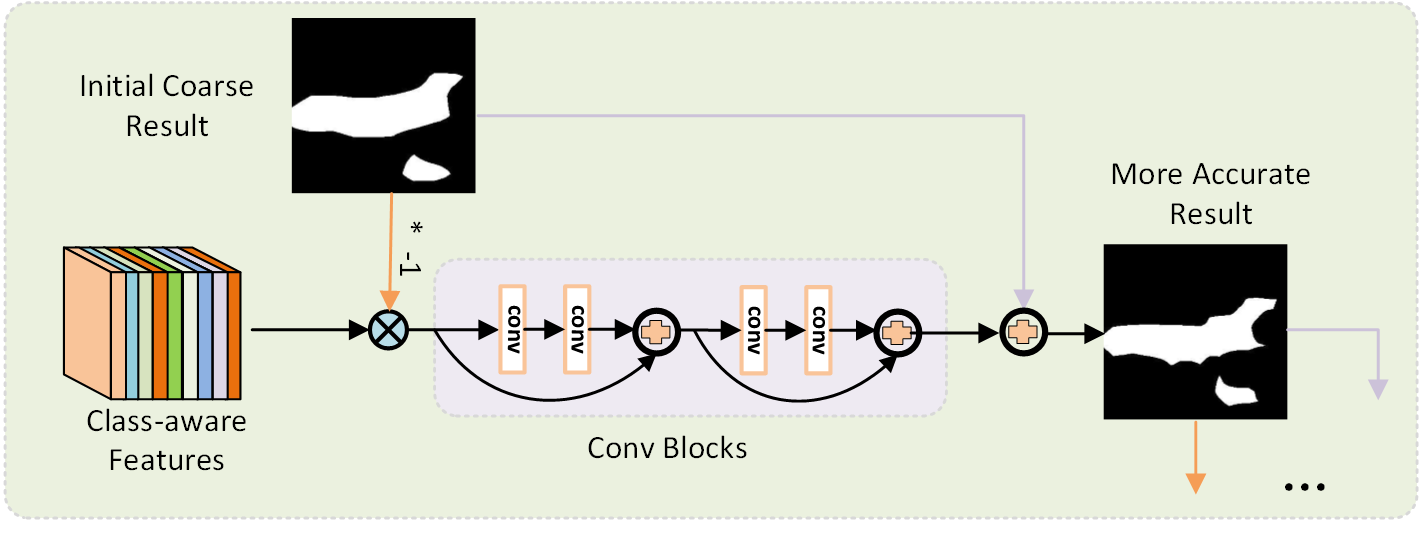}
	\end{center}
	\caption{The illustration of detail enrichment process. The initial coarse result would be leveraged as the earsing mask to force the network to perform the details mining.}
	\label{fig:detail}
\end{figure}

Though the DPMG refines the foreground area with the background information, the generated prior masks could only be leveraged as the guiding mask since the limited image resolution and the lack of semantic object details. The progressive semantic detail enrichment module aims at iteratively optimizing the parsing results by mining the hidden semantic details like object components and edges from the high-level layers to the low-level layers. The progressive optimizing process performs as follow: the query features are firstly fused with support prototypes(e.g., three prototypes generated by PMMs\cite{A29} ) by the prototype processing method in \cite{A29} to generate class-aware features at each level, which are simultaneously concatenated with the dual prior masks. Then we feed high-level class-aware features into the residual convolutional blocks to produce the initial coarse parsing results. Subsequently the initial coarse results are leveraged to erase the high-confidence foreground information in lower level class-aware feature, and a series of convolutional blocks are adopted to activate the semantic details, and the earsed high-confidence foreground area (initial coarse results) would be added with the semantic details to obtain the more accurate parsing results, which is directly supervised by the ground truth. This process is iteratively performed several times to obtain the parsing results with high resolution and semantic details. The detail enrichment process of one layer is shown in Figure \ref{fig:detail}.

Specifically, given the coarse parsing result ${R_I} = [R_P^I,R_N^I]$ , where ${R_P^I}$ denotes the foreground probability map, and ${R_N^I}$ denotes the background probability map, we adopt a softmax layer to get the normalized features ${\tilde R_I}=[{\tilde R_P^I},{\tilde R_N^I}]$, then we reverse the foreground probability ${\tilde R_P^I}$ to get the negative weight map ${E_n}$:
\begin{equation}
{E_n} =  - 1*{\tilde R_P^I} + 1
\end{equation} 
Besides, the element-wise multiplication between negative weight map and the lower layer class-aware features could readily erase the high-confidence foreground area of query images. Finally the processed class-aware features $F_{fused}$ which is fused with the support prototypes are propagated into a series of residual convolutional blocks to acquire the activated semantic details ${R_{II}} = [R_P^{II},R_N^{II}]$:
\begin{equation}
{R_{II}} = conv({F_{fused}} \odot {E_n})
\end{equation}

Moreover, the activated semantic details and initial coarse parsing result (high-confidence foreground area) are added to generate the optimized parsing result ${R_D}$: 
\begin{equation}
{R_D} = {R_I} + {R_{II}} = [R_P^I + R_P^{II},R_N^I + R_N^{II}]
\end{equation}
Each pixel of optimized parsing result ${R_D}$ has two corresponding possibility values. By comparing the probability  in pixel-level, we could predict the complete semantic objects ${R_C}$, which are supervised by the ground truths with cross-entropy loss function:
\begin{equation}
{R_C} = \mathop {\arg \max }\limits_i (soft\max ([R_P^D,R_N^D])),i \in [P,N]
\end{equation}

\begin{table*}[t!]
	\centering
	\scriptsize
	\footnotesize
	\renewcommand{\arraystretch}{1.1}
	\renewcommand{\tabcolsep}{5.8mm}		
	\caption{Class mIoU results on four folds of PASCAL-$5^i$. Our proposed model outperforms all previous methods under both VGG16 backbone and resnet50 backbone. We use \red{red} and \blu{blue} to indicate the two best scores.}
	\scalebox{0.8}{
		\begin{tabular}{cc|ccccc|ccccc}
			\hline
			\multicolumn{2}{c|}{\multirow{2}{*}{Method}}&  \multicolumn{5}{c|}{1-shot}&\multicolumn{5}{c}{5-shot} \\
			
			\multicolumn{2}{c|}{}     
			& split0  & Split1 & Split2 & Split3 & Mean  & split0  & Split1 & Split2 & Split3 & Mean \\ 
			\hline \hline
			\multicolumn{12}{c}{VGG16-Backbone}  \\ \hline  
			\multicolumn{2}{c|}{OSLSM\cite{A30}}   & 33.6 & 55.3 & 40.9 & 33.5 & 40.8 & 35.9 & 58.1 & 42.7 & 39.1 & 44.0 \\ 
			\multicolumn{2}{c|}{co-FCN\cite{A31}}    & 36.7 & 50.6 & 44.9 & 32.4 & 41.1 & 37.5 & 50.0 & 44.1 & 33.9 & 41.4\\ 
			\multicolumn{2}{c|}{SG-one\cite{A11}}    & 40.2 & 58.4 & 48.4 & 38.4 & 46.3 & 41.9 & 58.6 & 48.6 & 39.4 & 47.1\\ 
			\multicolumn{2}{c|}{AMP\cite{A13}}    & 41.9 & 50.2 & 46.7 & 34.7 & 43.4 & 41.8 & 55.5 & 50.3 & 39.9 & 46.9\\ 
			\multicolumn{2}{c|}{PANet\cite{A12}}   & 42.3 & 58.0 & 51.1 & 41.2 & 48.1 & 51.8 & 64.6 & \red{59.8} & 46.5 & 55.7\\ 
			\multicolumn{2}{c|}{FWB\cite{A32}}    & 47.0 & 59.6 & 52.6 & 48.3 & 51.9 & 50.9 & 62.9 & \blu{56.5} & 50.1 & 55.1\\ 
			\multicolumn{2}{c|}{RPMMs\cite{A29}}    & 47.1 & 65.8 & 50.6 & 48.5 & 53.0 & 50.0 & 66.5 & 51.9 & 47.6 & 54.0\\ 
			\multicolumn{2}{c|}{SST\cite{A68}}    &50.9 & 63.0 & 53.6 & 49.6 & 54.3 & 52.5 & 64.8 & 59.5 & 51.3 & 57.0\\ 
			\multicolumn{2}{c|}{ASR\cite{A64}}    &50.2 & 66.4 & 54.3 & 51.8 & 55.6 & 53.7 & 68.5 & 55.0 & \blu{54.8} & 57.9\\ 
			\multicolumn{2}{c|}{PFENet\cite{A33}}    & \blu{56.9} & \red{68.2} & \blu{54.4} & \red{52.4} & \blu{58.0} & \blu{59.0} & \red{69.1} & 54.8 & 52.9 & \blu{59.0}\\ 
			\multicolumn{2}{c|}{Ours}    & \red{61.4} & \blu{67.3} & \red{55.4} & \blu{52.0} & \red{59.1} & \red{64.4} & \blu{68.7} & \red{59.8} & \red{55.9} & \red{62.2}\\ 
			\hline
			\hline
			\multicolumn{12}{c}{Resnet50-Backbone}  \\ \hline
			\multicolumn{2}{c|}{CANet\cite{A14}}    & 52.5 & 65.9 & 51.3 & 51.9 & 55.4 & 55.5 & \blu{67.8} & 51.9 & 53.2 & 57.1\\ 
			\multicolumn{2}{c|}{PGNet\cite{A15}}    & 56.0 & 66.9 & 50.6 & 50.4 & 56.0 & 54.9 & 67.4 & 51.8 & 53.0 & 56.8\\ 
			\multicolumn{2}{c|}{PPNet\cite{A55}}    & 47.8 & 58.8 & 53.8 & 45.6 & 51.5 & 58.4 & 67.8 & \red{64.9} & 56.7 & \blu{62.0} \\ 
			\multicolumn{2}{c|}{RPMMs\cite{A29}}    & 55.2 & 66.9 & 52.6 & 50.6 & 56.3 & 56.3 & 67.3 & 54.5 & 51.0 & 57.3\\ 
			\multicolumn{2}{c|}{SST\cite{A68}}    &54.4 & 66.4 & \red{57.1} & 52.5 & 57.6 & 58.6 & 68.7 & 63.1 & 55.3 & 61.4\\ 
			\multicolumn{2}{c|}{DAN\cite{A67}}    &- & - & - & - & 57.1 & - & - & - & - & 59.5\\ 
			\multicolumn{2}{c|}{ASR\cite{A64}}    &55.2 & \red{70.4} & 53.4 & 53.7 & 58.2 & 59.4 & \red{71.8} & 56.9 & 55.7 & 61.0\\ 
			\multicolumn{2}{c|}{PFENet\cite{A33}}   & \blu{61.7} & 69.5 & 55.4 & \red{56.3} & \blu{60.8} & \blu{63.1} & \blu{70.7} & 55.8 & \blu{57.9} & 61.9\\ 
			\multicolumn{2}{c|}{Ours}    & \red{64.2} & \blu{69.0} & \blu{55.5} & \blu{55.5} & \red{61.1} & \red{67.8} & \blu{70.7} & \blu{58.9} & \red{60.0} & \red{64.4}\\ 
			\hline
			\hline
			\multicolumn{12}{c}{Resnet101-Backbone}  \\ \hline
			\multicolumn{2}{c|}{FWB\cite{A32}}    & 51.3 & 64.5 & \blu{56.7} & 52.2 & 56.2 & 54.8 & 67.4 & \red{62.2} & 55.3 & 59.9\\ 
			\multicolumn{2}{c|}{DAN\cite{A65}}    & 54.7 & 68.6 & \red{57.8} & 51.6 & 58.2 & 57.9 & 69.0 & \blu{60.1} & 54.9 & 60.5\\ 
			\multicolumn{2}{c|}{PFENet\cite{A33}}   & \blu{60.5} & \red{69.4} & 54.4 & \red{55.9} & \blu{60.1} & \blu{62.8} & \blu{70.4} & 54.9 & \blu{57.6} & \blu{61.4}\\ 
			\multicolumn{2}{c|}{Ours}    & \red{65.6} & \blu{69.0} & 54.4 & \blu{53.9} & \red{60.7} & \red{70.1} & \red{70.9} & 56.0 & \red{59.5} & \red{64.1}\\ 
			\hline
	\end{tabular}}
	
	\label{pascal}	
\end{table*}

\begin{table*}[t!]
	\centering
	\scriptsize
	\footnotesize
	\renewcommand{\arraystretch}{1.1}
	\renewcommand{\tabcolsep}{5.8mm}
	\caption{Class mIoU results on four folds of MS COCO. Our proposed method outperforms all previous methods under both VGG16 backbone and resnet50 backbone. FWBF and RPMMS adopt Resnet101 backbone  while others use resnet50 backbone. We use \red{red} and \blu{blue} to indicate the two best scores.}
	\scalebox{0.80}{
		\begin{tabular}{cc|ccccc|ccccc}
			\hline
			\multicolumn{2}{c|}{\multirow{2}{*}{Method}}&  \multicolumn{5}{c|}{1-shot}
			&\multicolumn{5}{c}{5-shot} \\
			\multicolumn{2}{c|}{}     
			& split0  & Split1 & Split2 & Split3 & Mean  & split0  & Split1 & Split2 & Split3 & Mean \\ 
			\hline
			\hline
			\multicolumn{12}{c}{VGG16-Backbone}   \\ \hline
			\multicolumn{2}{c|}{PANet\cite{A12}}    & - & - & - & - & 20.9 & - & - & - & - & 29.7\\
			\multicolumn{2}{c|}{FWB\cite{A32}}    & 18.4 & 16.7 & 19.6 & 25.4 & 20.0 & 20.9 & 19.2 & 21.9 & 28.4 & 22.6\\ 
			\multicolumn{2}{c|}{PFENet\cite{A28}}    & \blu{33.4} & \red{36.0} & \blu{34.1} & \blu{32.8} & \blu{34.1} & \blu{35.9} & \red{40.7} & \blu{38.1} & \blu{36.1} & \blu{37.7}\\ 
			\multicolumn{2}{c|}{Ours}    & \red{35.2} & \blu{35.3} & \red{35.5} & \red{33.4} & \red{34.8} & \red{42.4} & \blu{37.4} & \red{41.3} & \red{41.2} & \red{40.6}\\
			\hline
			\hline
			\multicolumn{12}{c}{Resnet50-Backbone}   \\ \hline
			\multicolumn{2}{c|}{PPNet\cite{A55}}    & 28.1 & 30.8 & 29.5 & 27.7 & 29.0 & \blu{39.0} & \red{40.8} & 37.1 & \blu{37.3} & \blu{38.5}\\ 
			\multicolumn{2}{c|}{RPMMs\cite{A29}}    & 29.5 & \blu{36.8} & 28.9 & 27.0 & 30.6 & 33.8 & 41.9 & 32.9 & 33.3 & 35.5\\ 
			\multicolumn{2}{c|}{SST\cite{A68}}    &- & - & - & - & 22.2 & - & - & - & - & 31.3\\ 
			\multicolumn{2}{c|}{DAN\cite{A67}}    &- & - & - & - & 24.4 & - & - & - & - & 29.6\\ 
			\multicolumn{2}{c|}{ASR\cite{A64}}    &30.6 & 36.7 & \blu{32.7} & \blu{35.4} & \blu{33.9} & 33.1 & 39.5 & 34.2 & 36.2 & 35.8\\ 
			\multicolumn{2}{c|}{PFENet\cite{A28}}    & \blu{34.3} & 33.0 & 32.3 & 30.1 & 32.4 & 38.5 & 38.6 & \blu{38.2} & 34.3 & 37.4\\ 
			\multicolumn{2}{c|}{Ours}    & \red{37.9} & \red{38.1} & \red{36.6} & \red{38.6} & \red{37.8} & \red{45.2} & \blu{40.2} & \red{41.1} & \red{43.9} & \red{42.6}\\ 
			\hline
			\hline
			\multicolumn{12}{c}{Resnet101-Backbone}   \\ \hline
			\multicolumn{2}{c|}{FWB\cite{A32}}    & 19.9 & 18.0 & 21.0 & 28.9 & 21.2 & 19.1 & 21.5 & 23.9 & 30.1 & 23.7\\  
			\multicolumn{2}{c|}{PANet\cite{A12}}    & - & - & - & - & 24.4 & - & - & - & - & 29.6\\
			\multicolumn{2}{c|}{RPMMs\cite{A29}}    & 29.3 & 34.8 & 27.1 & 27.3 & 29.6 & 33.0 & 40.6 & 30.1 & 33.3 & 34.3\\ 
			\multicolumn{2}{c|}{PFENet\cite{A28}}    & \blu{36.8} & \red{41.8} & \blu{38.7} & \blu{36.7} & \blu{38.5} & \blu{40.4} & \red{46.8} & \blu{43.2} & \blu{40.5} & \blu{42.7}\\ 
			\multicolumn{2}{c|}{Ours}    & \red{39.7} & \red{41.9} & \red{37.9} & \red{38.7} & \red{39.6} & \red{43.5} & \blu{42.9} & \red{46.7} & \red{45.6} & \red{44.7}\\ 
			\hline
	\end{tabular}}
	\setlength{\abovecaptionskip}{0pt}%
	\setlength{\belowcaptionskip}{10pt}%
	
	\label{coco}	
\end{table*}

By iteratively repeating the optimized process with hierarchical information from high-level to low-level, namely detail enrichment,  we ultimately get the most accurate segmentation result in high resolution at lowest level. In our designed experiments, the last four layers outputs of backbones are utilized in progressive semantic detail enrichment module to get four progressively optimized parsing results, where the number of layers are further studied in the following section, and all parsing results in a single PSDE module are simultaneously supervised by same ground truths. Therefore, the overall loss of our model is defined as:
\begin{equation}
{\ell _{all}} = \sum\limits_{i = 1}^n {\ell _{part}^i} 
\end{equation}
where the ${\ell _{part}^i}$ denotes the loss of $i$-th parsing result, and the n is the number of parsing layers, namely the number of iterative operation.

\section{Experiments}

\subsection{Implements Details}

\begin{figure*}[t]
	\begin{center}
		\includegraphics[width=1.0\linewidth]{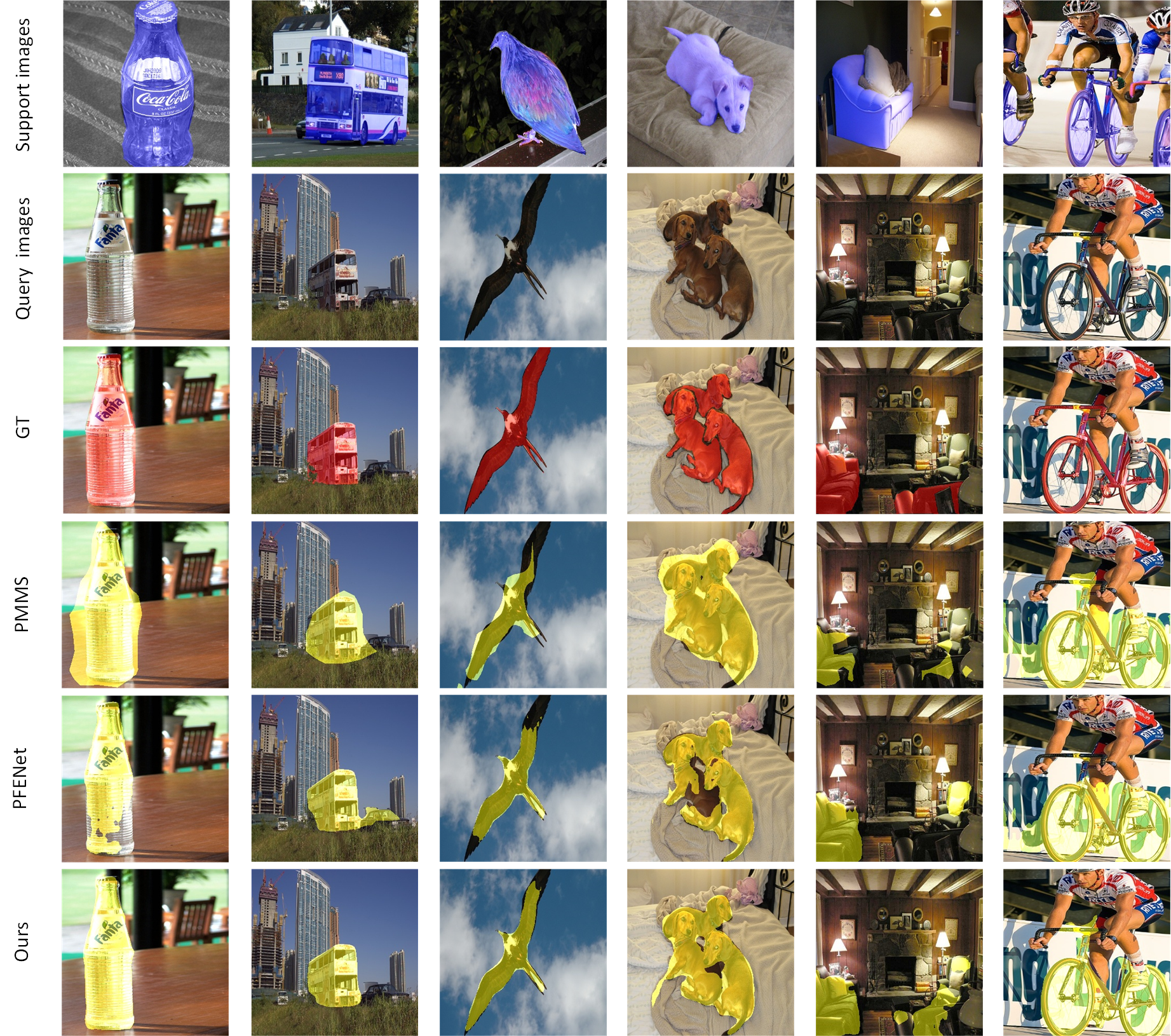}
	\end{center}
	\caption{ Qualitative results of our model. From top to bottom: support images, query images, ground truth of query images, predictions of PMMs network, predictions of PFENet, predictions of our network. The last two columns denote the false results. }
	\label{result}
\end{figure*}

\textbf{Datasets.} We follow the datasets splitting method in \cite{A29}, and testify our model in the PASCAL-$5^i$ dataset \cite{A30} and the MS COCO dataset \cite{A34}. PASCAL-$5^i$ is created by the combination of PASCAL VOC 2012 dataset\cite{A35} and the extended SDS dataset \cite{A36}. The 20 categories in PASCAL-$5^i$ are evenly divided into 4 splits, $i \in \left\{ {0,1,2,3} \right\}$. Subsequently, each split would has 5 classes. We randomly choose 3 splits for training, and the rest split for testing in a cross-validation manner. Similarly, The 80 category in MS COCO are also evenly divided into 4 splits. Thus each split has 20 classes. The classes in each split $i$ in could be written as $\{ 4i - 3 + j\} $, where $i \in \{ 1,2,....,20\} ,j \in \{ 0,1,2,3\} $. More details could be found in \cite{A37}. 3 splits would be randomly picked out for training and the rest split for testing. During testing, we follow the testing method in previous algorithms to randomly sample 1000 query-support pairs for evaluation.

\noindent \textbf{Experimental Setting.} Following the evaluation setting in \cite{A33}, we adopt the class mean intersection over union (mIOU) as our evaluation metric, which could straightly reflect the model performance. Formally, the mIOU could be defined as follow:
\begin{equation}
mIOU = \frac{1}{C}\sum\limits_{i = 1}^C {IO{U_i}}
\end{equation}
where $C$ is the number of categories in each split and ${IO{U_i}}$ is the  intersection over union of class $i$. The $C$ equals 20 for PASCAL-$5^i$ and equals 80 for MS COCO. 
We randomly sample 1,000 query-support pairs in each test. Our network is constructed on Pytorch. The VGG16 \cite{A2}, Resnet50 \cite{A3} and Resnet101 \cite{A3} networks pre-trained in Imagenet are adopted as backbones. Other layers are initialized by the default setting of PyTorch. Meanwhile, we utilize features of last four layers in backbones as the diverse inputs of our constructed network since surface features contain little semantic information and the number of input layers is further researched in ablation study. Model is trained by SGD optimizer with momentum of 0.9 for 50,000 iterations. The learning rate is 1e-3, and weight decay is 0.0005. The batchsize is 32. Our experiments are performed on an NVIDIA Titan Xp GPU. The input images are augmented with random horizontal flipping. In $k$-shot setting, the $k$-shot dual prior masks would be averaged to get the finally guiding dual prior mask.

\begin{table}
	\centering
	\scriptsize
	\footnotesize
	\renewcommand{\arraystretch}{1.0}
	\renewcommand{\tabcolsep}{5.2mm}
	\caption{Ablation study of dual prior mask generation module. The effect of foreground prior mask and the background prior mask are respectively studied.}
	\scalebox{0.8}{
		\begin{tabular}{ccc|c}
			\hline
			\multicolumn{1}{c}{\multirow{1}{*}{PSDE}}& 
			\multicolumn{1}{c}{Foreground Prior Mask} &\multicolumn{1}{c|}{Background Prior Mask} &\multicolumn{1}{c}{mIOU} \\
			\hline
			\multicolumn{1}{c}{\checkmark} & \multicolumn{2}{c|}{}  &58.5 \\ 
			\multicolumn{1}{c}{\checkmark} &\multicolumn{1}{c}{\checkmark} & \multicolumn{1}{c|}{}  &60.5  \\
			\multicolumn{1}{c}{\checkmark} &  \multicolumn{1}{c}{}&\multicolumn{1}{c|}
			{\checkmark}  &59.6 \\
			\multicolumn{1}{c}{\checkmark} & \multicolumn{1}{c}{\checkmark} &  \multicolumn{1}{c|}{\checkmark}  &\red{61.4}\\  \hline		
	\end{tabular}}
	
	\label{table1}
\end{table}
\begin{table}
	\centering
	\scriptsize
	\footnotesize
	\renewcommand{\arraystretch}{1.0}
	\renewcommand{\tabcolsep}{5.2mm}
	\caption{Ablation study of progressive semantic detail enrichment module. The Detail Enrichment means the designed detail mining process, and the Iterative Process denotes the we iteratively perform the details enrichment in diverse layers. }
	\scalebox{0.8}{
		\begin{tabular}{ccc|c}
			\hline
			\multicolumn{1}{c}{\multirow{1}{*}{DPMG}}&  \multicolumn{1}{c}{Detail Enrichment} &\multicolumn{1}{c|}{Iterative Process} &\multicolumn{1}{c}{mIOU} \\
			\hline
			\multicolumn{1}{c}{\checkmark} &    \multicolumn{2}{c|}{}  &58.2 \\ 
			\multicolumn{1}{c}{\checkmark} & \multicolumn{1}{c}{\checkmark} & \multicolumn{1}{c|}{}  &59.8    \\
			\multicolumn{1}{c}{\checkmark} &  \multicolumn{1}{c}{\checkmark} &  \multicolumn{1}{c|}{\checkmark}  &\red{61.4}\\  \hline		
	\end{tabular}}
	\label{table2}
\end{table}

\begin{figure*}
	\begin{center}
		\includegraphics[width=0.82\linewidth]{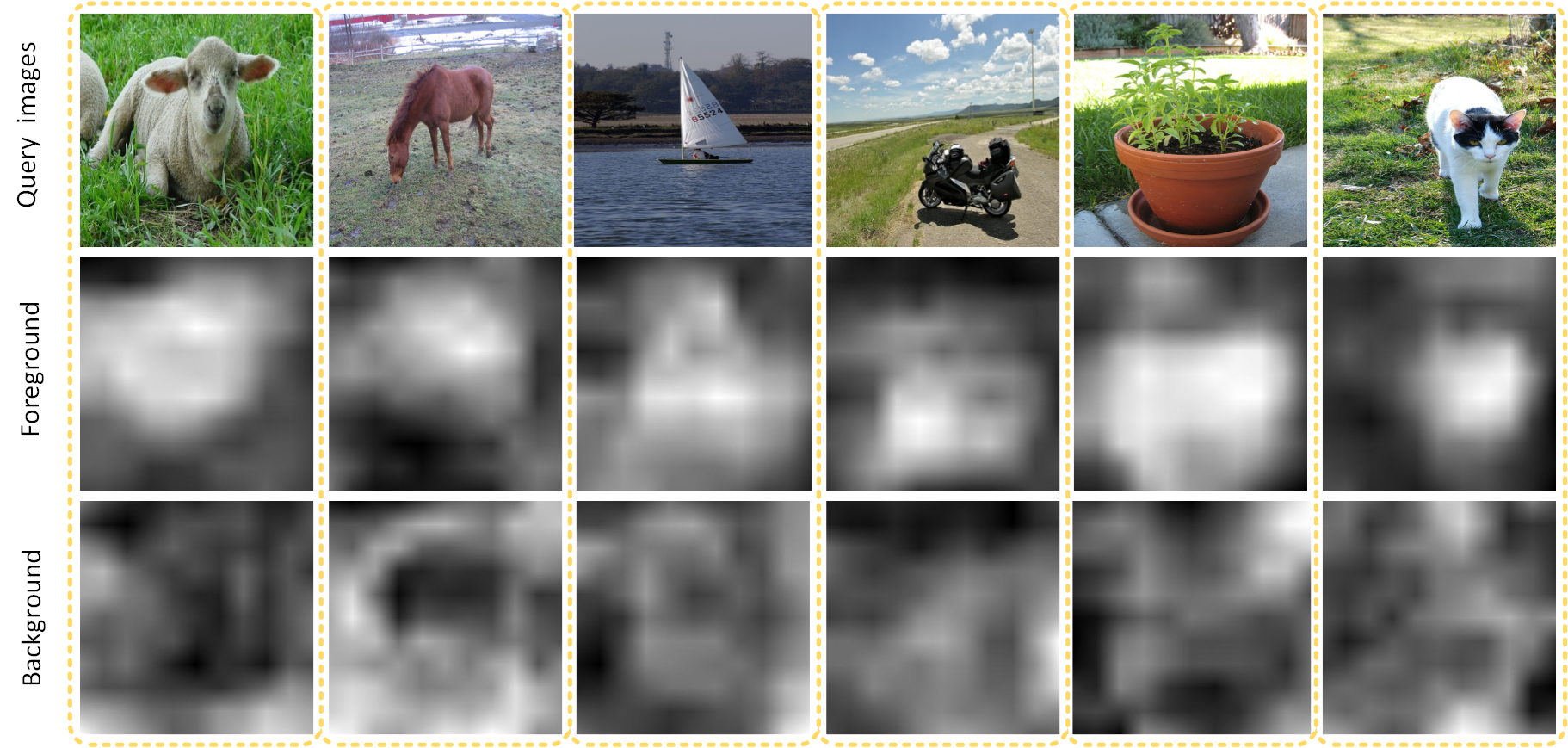}
	\end{center}
	\caption{Visualization of generated dual prior masks. Top row represents the query images, middle row denotes the foreground prior masks, and bottom row means the background prior masks. }
	\label{prior mask}
\end{figure*}

\subsection{Performance Analysis}
The comparison between our model and other methods in PASCAL-$5^i$  datasets is shown in Table \ref{pascal}. We could clearly find that our model outperforms all state-of-art models and gains better performance with the reinforcement of backbone. Particularly, with VGG16 backbone,  our model brings 1.1$\%$ mIoU improvement in 1-shot setting and increases mIoU by 3.2$\%$ in 5-shot setting. Besides, utilizing Resnet50 backbone, we could found that our model yields performance gain of  0.3$\%$ in 1-shot task and achieves 2.2$\%$ mIOU improvement in  5-shot setting. Moreover, With Resnet101 backbone, our network acquire 0.6$\%$ mIOU improvement in 1-shot setting and improve the performance with 2.7$\%$ gain in 5-shot setting.

As shown in Table \ref{coco}, our model still outperforms previous state-of-art algorithms of significant advantage in MS COCO dataset. With VGG16 backbone, the proposed method gets 0.7$\%$ mIOU improvement in 1-shot setting and 2.9$\%$ mIOU improvement in 5-shot setting. Moreover, by adopting the Resnet50 as backbone, our model outperforms the state-of-art methods with 3.9$\%$ performance gain in 1-shot setting and 4.1$\%$ performance gain in 5-shot setting. Furthermore, leveraging the Resnet101 as backbone, our network brings 1.1$\%$  mIOU improvement in 1-shot setting and yields performance gain of 2$\%$ in 5-shot setting.   If we scratch a little deep about the results in our Table \ref{coco}, we could find the results have larger performance gain compared with the results in PASCAL-$5^i$. We believe that the increasing of training data in MS COCO dataset could contribute to it since more abundant training data could help the network have stronger scene change adaptability and detail capturing ability. Another intriguing result is that although some methods utilize resnet101 as backbone which has more powerful ability of feature extraction, these models still fail to outperform the performance of our model, or even the results of their own model with VGG16 backbone. This suggests that superabundant parameters may exacerbate the performance in few-shot segmentation.

\begin{figure}
	\begin{center}
		\includegraphics[width=1.0\linewidth]{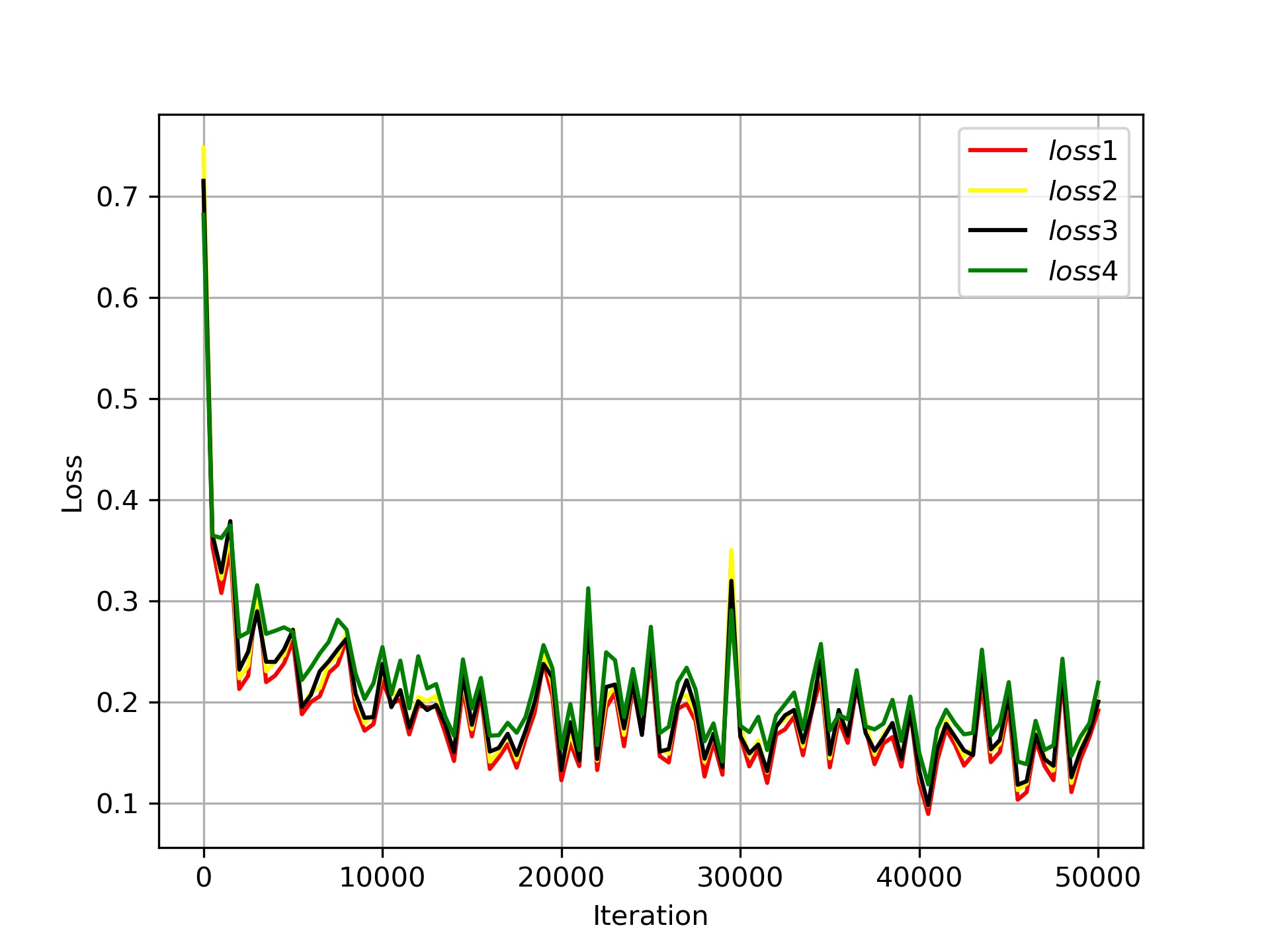}
	\end{center}
	\caption{Visualization of the loss curves corresponding to the outputs of diverse layers. \textbf{Loss1}:the loss curve for the outputs of high-level layers. \textbf{Loss2 and loss3}:the loss curve for the outputs of middle-level layers. \textbf{Loss4}:the loss curve for the outputs of low-level layers.}
	\label{loss}
\end{figure}

Meanwhile, we find that our model get extraordinary performance in split0 (aeroplane, bicycle, bird, boat, bottle) of PASCAL-$5^i$ datasets, After analyzing the split in PASCAL-$5^i$  dataset, we figure out that these classes have similar background between diverse images and relatively regular appearances. The similar background information could contribute to the location of foreground in query images, and the objects with regular appearances could be easier for network to mine the neglected object components and missing edge areas.

\begin{figure*}
	\begin{center}
		\includegraphics[width=0.80\linewidth]{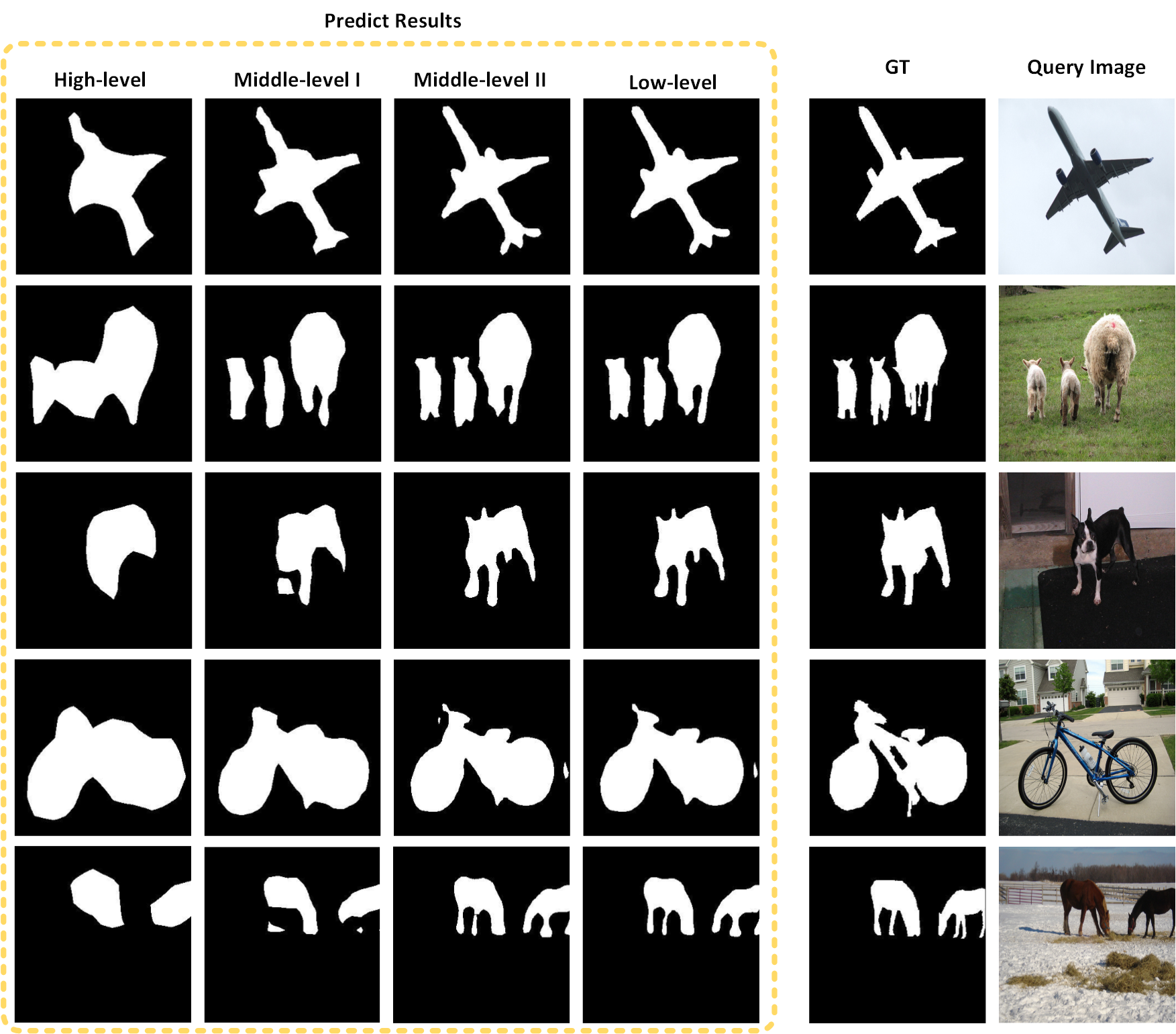}
	\end{center}
	\caption{Visualization of progressively details enrichment process. The left lines represents the predict results of diverse layers from the high-level to low-level. The right lines respectively denotes the query images and the corresponding ground truth.   }
	\label{details}
\end{figure*}

However, our model does not get the best performance in some splits like the split2 (potted plant, sheep, sofa, train, tv/monitor) of PASCAL-$5^i$ datasets. After analyzing the images of these splits, we figure out that some classes in the  splits have very tiny background areas like the potted plant class or only part structure of objects are shown in images like the train class. The tiny background could not provide enough comparable information for PSDE module and the part structure could result in the semantic incompletion which hampers the semantic detail mining process. The two issues clearly degrade the efficiency of DPMG and PSDE modules. It is obvious that this dilemma is alleviated in 5-shot setting. A reasonable explanation is that the uncertainty of background in some classes is decreased and the semantic structure information is further supplemented with the increasing of support images.

To better illustrate the functional effect of DPMG, some generated dual prior masks are shown in Figure \ref{prior mask}. The visualization of the dual prior masks clearly demonstrates that DPMG architecture could offer a better location of foreground in pixel-level.  Particularly, the most  class discriminative region are highlighted in the foreground prior masks. For the background prior masks, though the complexity of background would influence the activation area, the values of high-confidence foreground area remains at low-level. For instance, for the second line from the right, we could nearly see the outline of the horse in the background prior mask and the high-confidence foreground area of horse are highlighted in the foreground prior mask. Furthermore, for the third line from the right, the outline of the ship could be explicitly observed in the foregound prior mask. 

Moreover, to further study the function of the PSDE module, the training loss curves for the outputs of diverse layers are shown in the Figure \ref{loss}.  Apart from the beginning phase of the training process, the loss curves for the outputs of lower layers always remain under the loss curves for the outputs of higher layers. This observation exactly match the original design of the PSDE module and the network could get the best segmentation result in the lowest layer with high resolution.  Aiming at futher directly analyzing the progressively detail enrichment process, some predict results are illustrated in the Figure \ref{details}. It is obvious that the high-level layers could only predict coarse appearances for semantic objects. By iteratively perform the detail enrichment operation from high-level to lwo-level, the details of semantic objects are progressively added into the predict results. For example, the wings of the airplane are dynamically refined, the legs of these sheeps are iteratively detailed and the architecture of the bicycle is gradually figured out. These detail enrichment process all helps the model get better predicted results. Interestingly, the largest change of predict results exists between the high-level layers and the middle-level I, which could also be viewed in the Figure \ref{loss}, i.e., the gap between the loss1 and the loss2. This phenomenon indicates that the first detail enrichment operation contributes more around the whole PSDE module.  
To summarize, we could conclude that the designed DPMG and PSDE modules truly help the network to acquire more accurate predicted masks.

The qualitative results are shown in Figure \ref{result}. The satisfactory segmentation results demonstrate the strong generalization and details capture ability of our designed architecture. Specifically, the designed model could capture more edge information in bird class and more components in bottle class, and the confused background areas are clearly erased in bus class. Moreover, the complete foreground regions are readily parsed with limited error background in dog class. However, we could find that some background regions are misunderstood as foreground in bicycle class. We believe the strong positive relations between misunderstood background and true foreground could contribute to it. Meanwhile, it is obvious that there are some isolate wrong foreground parts in sofa class. The issue could be explained by the reason that the parsing model is forced to capture the details of semantic objects, which could result in the wrong parsing of tiny regions with rich semantic information.

\begin{table}[t]
	\centering
	\scriptsize
	\footnotesize
	\renewcommand{\arraystretch}{1.0}
	\renewcommand{\tabcolsep}{3.6mm}
	\caption{ Ablation study on the number of generated prototypes.}
	\scalebox{1.0}{
		\begin{tabular}{c|cccc}
			\hline
			\multicolumn{1}{c|}{Numbers}&  \multicolumn{1}{c}{2} &\multicolumn{1}{c}{3} &\multicolumn{1}{c}{4} &\multicolumn{1}{c}{5} \\
			\hline
			\multicolumn{1}{c|}{mIOU} & 59.8 &  \red{61.4}  & 61.3 & 60.8 \\
			\hline	
	\end{tabular} }
	
	\label{table3}
\end{table}

\begin{table}[t]
	\centering
	\scriptsize
	\footnotesize
	\renewcommand{\arraystretch}{1.0}
	\renewcommand{\tabcolsep}{3.6mm}
	\caption{ Ablation study on the number of enrichment layers.}
	\scalebox{1.0}{
		\begin{tabular}{c|cccc}
			\hline
			\multicolumn{1}{c|}{Numbers}&  \multicolumn{1}{c}{2} &\multicolumn{1}{c}{3} &\multicolumn{1}{c}{4} &\multicolumn{1}{c}{5} \\
			\hline
			\multicolumn{1}{c|}{mIOU} & 60.0 &  61.0  & \red{61.4} & 61.3 \\
			\hline	
	\end{tabular} }
	
	\label{table4}
\end{table}

\subsection{Ablation study}

In order to demonstrate the efficiency of our designed modules, we set some ablation experiments in split-0 of PASCAL-$5^i$ with 1-shot setting and VGG-16 backbone.

Dual prior mask generation module aims at eliminating the wrong activated background regions by supplementing background area as refinable scene information. As shown in Table \ref{table1}, our model gets 2.9$\%$ mIOU improvement with dual prior masks. The introduction of the foreground  prior mask help our model acquire 2.0 $\%$ and the utilization of background prior mask would  obtain 1.1 $\%$ mIOU improvement.  From these results, we could figure out that the DPMG module could offer a steady understanding of query images for following parsing module. 

\begin{figure}[t]
	\begin{center}
		\includegraphics[width=1.0\linewidth]{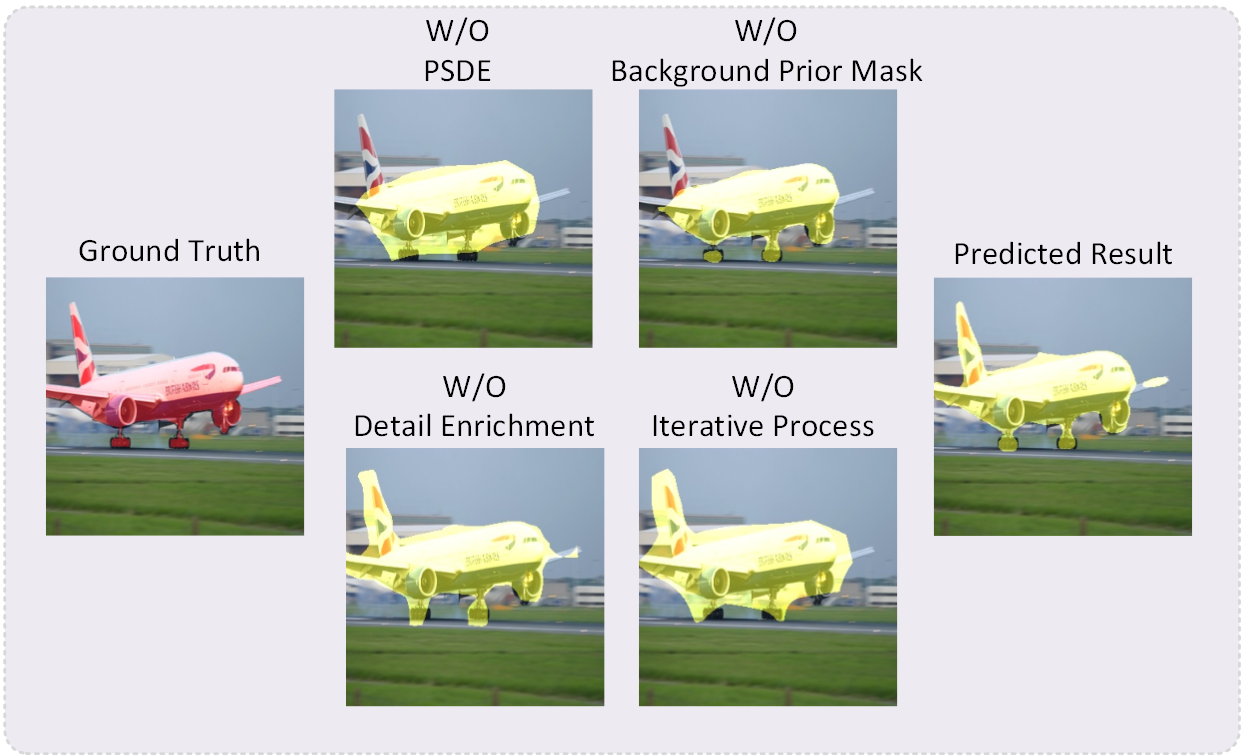}
	\end{center}
	\caption{Qualitative results of ablation study. \textbf{W/O}:without.\textbf{W/O PSDE} denotes the designed algorithm without PSDE module,  \textbf{W/O background prior mask} denotes the designed algorithm without background prior mask, \textbf{W/O detail enrichment} means that we directly combine the features from the different layers to predict results without the details mining process, and the  \textbf{W/O iterative process} illustrates that we only perform the details enrichment on single layer.  }
	\label{ablation}
\end{figure}

The results in Table \ref{table2} further demonstrate the ability of semantic detail enrichment module. The combination of iteratively foreground erasing process and detail enrichment between neighbor layers constitutes the semantic details enrichment module. The fusion of iterative operation and detail enrichment provides 3.2$\%$ mIOU improvement. The detail enrichment helps network forget the high-confidence foreground area and focus on precisely parsing the details of semantic objects. Thus the single detail enrichment helps model gain 1.6$\%$ performance improvement. Meanwhile, iterative operation could maximally utilize the semantic features from different levels to predict the most accurate the parsing result with high resolution, and it helps the network get 1.6$\%$ performance improvement compared with single detail enrichment.  Based on above findings, the PSDE module clearly demostrate its strong detail enrichment ability.

Moveover, some experiments are performed to analyze the effect of the number of generated prototypes, and the experiments results are shown in the Table \ref{table3}. It is clear that the mIOU performance gets the best performance when the number of generated prototypes equals 3. This observation tells us that modelling the semantic objects with multiple prototypes truly has upper bound. Therefore, the number of generated prototypes is set as 3 for other experiments. 

The ablation study on the number of enrichment layers are shown in the Table \ref{table4}. The performance of our model has an improvement with the increasing of number of enrichment layers till number equals 4. We could figure out that the model gets best performance, i.e., 61.4 mIOU, when the number equals 4 and the model captures the max details at the same time.  Thus the number of enrichment layers is set as 4 for other experiments.

Finally, to further directly view the function of our designed architecture, the qualitative results of ablation study are shown in Figure \ref{ablation}. It is obvious that the segmentation result would be coarse without PSDE module. Meanwhile, the lack of background prior mask would result in the missing of some foreground details like object components and edge areas. An interesting result is that although the detail enrichment and iterative process both benefit the parsing result, the foreground erasing has better performance and the detail enrichment has the ability to capture  abundant semantic details like components or edge areas, which could be figured out by the comparison of diverse parsing results.

\section{Conclusion}

In this paper, we propose a novel few-shot semantic segmentation with dual prior mask generation module and progressive semantic detail enrichment module, which elaborately guide the training model to be scene-adaptive and detail-attentive. The dual prior mask generation module would produce dual prior masks to refine the location of foreground. The progressive detail enrichment module progressively modifies the predicted mask by the fusion of iteratively foreground erasing and detail enrichment in a hierarchical manner, which precisely grasps all details of semantic objects in diverse scales. Extensive experiments are elaborately conducted on the challenging PASCAL-$5^i$ and MS COCO datasets, and we achieve great performances in both two datasets.




%





\ifCLASSOPTIONcaptionsoff
  \newpage
\fi

\vfill



\begin{thebibliography}{29}

	
\bibitem{A18}
Vijay Badrinarayanan, Alex Kendall, and Roberto Cipolla.
\newblock Segnet: A deep convolutional encoder-decoder architecture for image
segmentation.
\newblock {\em IEEE Transactions on Pattern Analysis and Machine Intelligence},
39(12):2481--2495, 2017.

\bibitem{A40}
Wonmin Byeon, Thomas~M Breuel, Federico Raue, and Marcus Liwicki.
\newblock Scene labeling with lstm recurrent neural networks.
\newblock In {\em Proceedings of the IEEE Conference on Computer Vision and
	Pattern Recognition}, pages 3547--3555, 2015.

\bibitem{A9}
Liang-Chieh Chen, George Papandreou, Iasonas Kokkinos, Kevin Murphy, and Alan~L
Yuille.
\newblock Deeplab: Semantic image segmentation with deep convolutional nets,
atrous convolution, and fully connected crfs.
\newblock {\em IEEE Transactions on Pattern Analysis and Machine Intelligence},
40(4):834--848, 2017.

\bibitem{A46}
Liang-Chieh Chen, George Papandreou, Florian Schroff, and Hartwig Adam.
\newblock Rethinking atrous convolution for semantic image segmentation.
\newblock {\em arXiv preprint arXiv:1706.05587}, 2017.

\bibitem{A47}
Liang-Chieh Chen, Yukun Zhu, George Papandreou, Florian Schroff, and Hartwig
Adam.
\newblock Encoder-decoder with atrous separable convolution for semantic image
segmentation.
\newblock In {\em Proceedings of the European Conference on Computer Vision
	(ECCV)}, pages 801--818, 2018.

\bibitem{A20}
Yunpeng Chen, Marcus Rohrbach, Zhicheng Yan, Yan Shuicheng, Jiashi Feng, and
Yannis Kalantidis.
\newblock Graph-based global reasoning networks.
\newblock In {\em Proceedings of the IEEE Conference on Computer Vision and
	Pattern Recognition}, pages 433--442, 2019.

\bibitem{A53}
Nanqing Dong and Eric~P Xing.
\newblock Few-shot semantic segmentation with prototype learning.
\newblock In {\em BMVC}, volume~3, 2018.

\bibitem{A35}
Mark Everingham, Luc Van~Gool, Christopher~KI Williams, John Winn, and Andrew
Zisserman.
\newblock The pascal visual object classes (voc) challenge.
\newblock {\em International Journal of Computer Vision}, 88(2):303--338, 2010.

\bibitem{A42}
Heng Fan and Haibin Ling.
\newblock Dense recurrent neural networks for scene labeling.
\newblock {\em arXiv preprint arXiv:1801.06831}, 2018.

\bibitem{A49}
Damien Fourure, R{\'e}mi Emonet, Elisa Fromont, Damien Muselet, Alain Tremeau,
and Christian Wolf.
\newblock Residual conv-deconv grid network for semantic segmentation.
\newblock {\em arXiv preprint arXiv:1707.07958}, 2017.

\bibitem{A45}
Jun Fu, Jing Liu, Haijie Tian, Yong Li, Yongjun Bao, Zhiwei Fang, and Hanqing
Lu.
\newblock Dual attention network for scene segmentation.
\newblock In {\em Proceedings of the IEEE/CVF Conference on Computer Vision and
	Pattern Recognition}, pages 3146--3154, 2019.

\bibitem{A51}
Jun Fu, Jing Liu, Yuhang Wang, Jin Zhou, Changyong Wang, and Hanqing Lu.
\newblock Stacked deconvolutional network for semantic segmentation.
\newblock {\em IEEE Transactions on Image Processing}, 2019.

\bibitem{A36}
Bharath Hariharan, Pablo Arbel{\'a}ez, Lubomir Bourdev, Subhransu Maji, and
Jitendra Malik.
\newblock Semantic contours from inverse detectors.
\newblock In {\em Proceedings of the IEEE International Conference on Computer
	Vision}, pages 991--998, 2011.

\bibitem{A6}
Kaiming He, Georgia Gkioxari, Piotr Doll{\'a}r, and Ross Girshick.
\newblock Mask r-cnn.
\newblock In {\em Proceedings of the IEEE International Conference on Computer
	Vision}, pages 2961--2969, 2017.

\bibitem{A3}
Kaiming He, Xiangyu Zhang, Shaoqing Ren, and Jian Sun.
\newblock Deep residual learning for image recognition.
\newblock In {\em Proceedings of the IEEE Conference on Computer Vision and
	Pattern Recognition}, pages 770--778, 2016.

\bibitem{A60}
Tao Hu, Pengwan Yang, Chiliang Zhang, Gang Yu, Yadong Mu, and Cees~GM Snoek.
\newblock Attention-based multi-context guiding for few-shot semantic
segmentation.
\newblock In {\em Proceedings of the AAAI Conference on Artificial
	Intelligence}, volume~33, pages 8441--8448, 2019.

\bibitem{A37}
Tao Hu, Pengwan Yang, Chiliang Zhang, Gang Yu, Yadong Mu, and Cees~GM Snoek.
\newblock Attention-based multi-context guiding for few-shot semantic
segmentation.
\newblock In {\em Proceedings of the AAAI Conference on Artificial
	Intelligence}, volume~33, pages 8441--8448, 2019.

\bibitem{A50}
Alex Kendall, Vijay Badrinarayanan, and Roberto Cipolla.
\newblock Bayesian segnet: Model uncertainty in deep convolutional
encoder-decoder architectures for scene understanding.
\newblock {\em arXiv preprint arXiv:1511.02680}, 2015.

\bibitem{A1}
Alex Krizhevsky, Ilya Sutskever, and Geoffrey~E Hinton.
\newblock Imagenet classification with deep convolutional neural networks.
\newblock {\em Advances in Neural Information Processing Systems},
25:1097--1105, 2012.

\bibitem{A56}
Gen Li, Varun Jampani, Laura Sevilla-Lara, Deqing Sun, Jonghyun Kim, and
Joongkyu Kim.
\newblock Adaptive prototype learning and allocation for few-shot segmentation.
\newblock {\em arXiv preprint arXiv:2104.01893}, 2021.

\bibitem{A7}
Guosheng Lin, Anton Milan, Chunhua Shen, and Ian Reid.
\newblock Refinenet: Multi-path refinement networks for high-resolution
semantic segmentation.
\newblock In {\em Proceedings of the IEEE Conference on Computer Vision and
	Pattern Recognition}, pages 1925--1934, 2017.

\bibitem{A34}
Tsung-Yi Lin, Michael Maire, Serge Belongie, James Hays, Pietro Perona, Deva
Ramanan, Piotr Doll{\'a}r, and C~Lawrence Zitnick.
\newblock Microsoft coco: Common objects in context.
\newblock In {\em European Conference on Computer Vision}, pages 740--755.
Springer, 2014.

\bibitem{A55}
Yongfei Liu, Xiangyi Zhang, Songyang Zhang, and Xuming He.
\newblock Part-aware prototype network for few-shot semantic segmentation.
\newblock In {\em European Conference on Computer Vision}, pages 142--158.
Springer, 2020.

\bibitem{A8}
Jonathan Long, Evan Shelhamer, and Trevor Darrell.
\newblock Fully convolutional networks for semantic segmentation.
\newblock In {\em Proceedings of the IEEE Conference on Computer Vision and
	Pattern Recognition}, pages 3431--3440, 2015.

\bibitem{A32}
Khoi Nguyen and Sinisa Todorovic.
\newblock Feature weighting and boosting for few-shot segmentation.
\newblock In {\em Proceedings of the IEEE International Conference on Computer
	Vision}, pages 622--631, 2019.

\bibitem{A48}
Hyeonwoo Noh, Seunghoon Hong, and Bohyung Han.
\newblock Learning deconvolution network for semantic segmentation.
\newblock In {\em Proceedings of the IEEE International Conference on Computer
	Vision}, pages 1520--1528, 2015.

\bibitem{A39}
Pedro Pinheiro and Ronan Collobert.
\newblock Recurrent convolutional neural networks for scene labeling.
\newblock In {\em International Conference on Machine Learning}, pages 82--90.
PMLR, 2014.

\bibitem{A44}
Tobias Pohlen, Alexander Hermans, Markus Mathias, and Bastian Leibe.
\newblock Full-resolution residual networks for semantic segmentation in street
scenes.
\newblock In {\em Proceedings of the IEEE Conference on Computer Vision and
	Pattern Recognition}, pages 4151--4160, 2017.

\bibitem{A31}
Kate Rakelly, Evan Shelhamer, Trevor Darrell, Alyosha Efros, and Sergey Levine.
\newblock Conditional networks for few-shot semantic segmentation.
\newblock {\em ICLR Workshop}, 2018.

\bibitem{A5}
Joseph Redmon, Santosh Divvala, Ross Girshick, and Ali Farhadi.
\newblock You only look once: Unified, real-time object detection.
\newblock In {\em Proceedings of the IEEE Conference on Computer Vision and
	Pattern Recognition}, pages 779--788, 2016.

\bibitem{A4}
Shaoqing Ren, Kaiming He, Ross Girshick, and Jian Sun.
\newblock Faster r-cnn: towards real-time object detection with region proposal
networks.
\newblock {\em IEEE Transactions on Pattern Analysis and Machine intelligence},
39(6):1137--1149, 2016.

\bibitem{A10}
Olga Russakovsky, Jia Deng, Hao Su, Jonathan Krause, Sanjeev Satheesh, Sean Ma,
Zhiheng Huang, Andrej Karpathy, Aditya Khosla, Michael Bernstein, et~al.
\newblock Imagenet large scale visual recognition challenge.
\newblock {\em International Journal of Computer Vision}, 115(3):211--252,
2015.

\bibitem{A30}
Amirreza Shaban, Shray Bansal, Zhen Liu, Irfan Essa, and Byron Boots.
\newblock One-shot learning for semantic segmentation.
\newblock {\em arXiv preprint arXiv:1709.03410}, 2017.

\bibitem{A52}
Bing Shuai, Zhen Zuo, Bing Wang, and Gang Wang.
\newblock Dag-recurrent neural networks for scene labeling.
\newblock In {\em Proceedings of the IEEE Conference on Computer Vision and
	Pattern Recognition}, pages 3620--3629, 2016.

\bibitem{A13}
Mennatullah Siam, Boris~N Oreshkin, and Martin Jagersand.
\newblock Amp: Adaptive masked proxies for few-shot segmentation.
\newblock In {\em Proceedings of the IEEE International Conference on Computer
	Vision}, pages 5249--5258, 2019.

\bibitem{A2}
Karen Simonyan and Andrew Zisserman.
\newblock Very deep convolutional networks for large-scale image recognition.
\newblock {\em arXiv preprint arXiv:1409.1556}, 2014.

\bibitem{A33}
Zhuotao Tian, Hengshuang Zhao, Michelle Shu, Zhicheng Yang, Ruiyu Li, and Jiaya
Jia.
\newblock Prior guided feature enrichment network for few-shot segmentation.
\newblock {\em IEEE Annals of the History of Computing}, (01):1--1, 2020.

\bibitem{A43}
Francesco Visin, Marco Ciccone, Adriana Romero, Kyle Kastner, Kyunghyun Cho,
Yoshua Bengio, Matteo Matteucci, and Aaron Courville.
\newblock Reseg: A recurrent neural network-based model for semantic
segmentation.
\newblock In {\em Proceedings of the IEEE Conference on Computer Vision and
	Pattern Recognition Workshops}, pages 41--48, 2016.

\bibitem{A12}
Kaixin Wang, Jun~Hao Liew, Yingtian Zou, Daquan Zhou, and Jiashi Feng.
\newblock Panet: Few-shot image semantic segmentation with prototype alignment.
\newblock In {\em Proceedings of the IEEE International Conference on Computer
	Vision}, pages 9197--9206, 2019.

\bibitem{A28}
Hang Xu, ChenHan Jiang, Xiaodan Liang, Liang Lin, and Zhenguo Li.
\newblock Reasoning-rcnn: Unifying adaptive global reasoning into large-scale
object detection.
\newblock In {\em Proceedings of the IEEE Conference on Computer Vision and
	Pattern Recognition}, pages 6419--6428, 2019.

\bibitem{A29}
Boyu Yang, Chang Liu, Bohao Li, Jianbin Jiao, and Qixiang Ye.
\newblock Prototype mixture models for few-shot semantic segmentation.
\newblock In {\em European Conference on Computer Vision}, pages 763--778.
Springer, 2020.

\bibitem{A16}
Fisher Yu and Vladlen Koltun.
\newblock Multi-scale context aggregation by dilated convolutions.
\newblock {\em arXiv preprint arXiv:1511.07122}, 2015.

\bibitem{A15}
Chi Zhang, Guosheng Lin, Fayao Liu, Jiushuang Guo, Qingyao Wu, and Rui Yao.
\newblock Pyramid graph networks with connection attentions for region-based
one-shot semantic segmentation.
\newblock In {\em Proceedings of the IEEE International Conference on Computer
	Vision}, pages 9587--9595, 2019.

\bibitem{A14}
Chi Zhang, Guosheng Lin, Fayao Liu, Rui Yao, and Chunhua Shen.
\newblock Canet: Class-agnostic segmentation networks with iterative refinement
and attentive few-shot learning.
\newblock In {\em Proceedings of the IEEE Conference on Computer Vision and
	Pattern Recognition}, pages 5217--5226, 2019.

\bibitem{A11}
Xiaolin Zhang, Yunchao Wei, Yi Yang, and Thomas~S Huang.
\newblock Sg-one: Similarity guidance network for one-shot semantic
segmentation.
\newblock {\em IEEE Transactions on Cybernetics}, 50(9):3855--3865, 2020.

\bibitem{A17}
Hengshuang Zhao, Jianping Shi, Xiaojuan Qi, Xiaogang Wang, and Jiaya Jia.
\newblock Pyramid scene parsing network.
\newblock In {\em Proceedings of the IEEE Conference on Computer Vision and Pattern Recognition}, pages 2881--2890, 2017.

\bibitem{A57}
Hyeonwoo Noh, Seunghoon Hong, and Bohyung Han.
\newblock Learning deconvolution network for semantic segmentation.
\newblock In {\em Proceedings of the IEEE International Conference on Computer	Vision}, pages 1520--1528, 2015.
\bibitem{A58}
Nicolas Carion, Francisco Massa, Gabriel Synnaeve, Nicolas Usunier, Alexander
Kirillov, and Sergey Zagoruyko.
\newblock End-to-end object detection with transformers.
\newblock In {\em European Conference on Computer Vision}, pages 213--229.
Springer, 2020.
\bibitem{A59}
Tsung-Han Chan, Kui Jia, Shenghua Gao, Jiwen Lu, Zinan Zeng, and Yi Ma.
\newblock Pcanet: A simple deep learning baseline for image classification?
\newblock {\em IEEE Transactions on Image Processing}, 24(12):5017--5032, 2015.

\bibitem{A62}
Guo-Sen Xie, Jie Liu, Huan Xiong, and Ling Shao.
\newblock Scale-aware graph neural network for few-shot semantic segmentation.
\newblock In {\em Proceedings of the IEEE Conference on Computer Vision and Pattern Recognition}, pages 5475--5484, 2021.
\bibitem{A63}
Boyu Yang, Fang Wan, Chang Liu, Bohao Li, Xiangyang Ji, and Qixiang Ye.
\newblock Part-based semantic transform for few-shot semantic segmentation.
\newblock {\em IEEE Transactions on Neural Networks and Learning Systems},
2021.
\bibitem{A64}
Binghao Liu, Yao Ding, Jianbin Jiao, Xiangyang Ji, and Qixiang Ye.
\newblock Anti-aliasing semantic reconstruction for few-shot semantic
segmentation.
\newblock In {\em Proceedings of the IEEE/CVF Conference on Computer Vision and Pattern Recognition}, pages 9747--9756, 2021.
\bibitem{A65}
Haochen Wang, Xudong Zhang, Yutao Hu, Yandan Yang, Xianbin Cao, and Xiantong
Zhen.
\newblock Few-shot semantic segmentation with democratic attention networks.
\newblock In {\em Computer Vision--ECCV 2020: 16th European Conference,
	Glasgow, UK, August 23--28, 2020, Proceedings, Part XIII 16}, pages 730--746.
Springer, 2020.
\bibitem{A66}
Huajun Liu, Fuqiang Liu, Xinyi Fan, and Dong Huang.
\newblock Polarized self-attention: Towards high-quality pixel-wise regression.
\newblock {\em arXiv preprint arXiv:2107.00782}, 2021.

\bibitem{A67}
Haochen Wang, Xudong Zhang, Yutao Hu, Yandan Yang, Xianbin Cao, and Xiantong
Zhen.
\newblock Few-shot semantic segmentation with democratic attention networks.
\newblock In {\em Computer Vision--ECCV 2020: 16th European Conference,
	Glasgow, UK, August 23--28, 2020, Proceedings, Part XIII 16}, pages 730--746.
Springer, 2020.

\bibitem{A68}
Kai Zhu, Wei Zhai, Zheng-Jun Zha, and Yang Cao.
\newblock Self-supervised tuning for few-shot segmentation.
\newblock {\em arXiv preprint arXiv:2004.05538}, 2020.

\end{thebibliography}
\end{document}